\documentclass{article}
\usepackage[hyphens]{url}
\usepackage[colorlinks=true,linkcolor=black,anchorcolor=black,citecolor=black,filecolor=black,menucolor=black,runcolor=black,urlcolor=black]{hyperref}
\usepackage{spconf,graphicx,tabularx}
\usepackage[noadjust]{cite}
\usepackage{enumerate}
\usepackage{amsmath,amsfonts}
\usepackage{bbm}
\usepackage{array}
\usepackage{scalerel} 
\usepackage{tikz}
\usetikzlibrary{calc,shapes,positioning,svg.path}
\newcommand{\token}[1]{$\stretchto[1200]{\langle}{10pt}${\texttt{\small#1}}$\stretchto[1200]{\rangle}{10pt}$}

\usepackage{xcolor}

\usepackage{tikz}
\newcommand*\circled[1]{%
  \begin{tikzpicture}
    \node[draw,circle,inner sep=1pt] {#1};
  \end{tikzpicture}}

\def\df{d_\mathrm{f}}
\def\dh{d_\mathrm{h}}
\def\da{d_\mathrm{a}}

\def\dpf{d_\mathrm{p}}
\def\le{l_\mathrm{e}}
\def\ld{l_\mathrm{d}}
\def\Thetae{\Theta_\mathrm{e}}
\def\Thetad{\Theta_\mathrm{d}}

\definecolor{darkgreen}{rgb}{0,0.7,0}
\definecolor{darkred}{rgb}{0.8,0.0,0}
\colorlet{redish}{darkred!60!white}

\DeclareMathOperator{\Enc}{Enc}
\DeclareMathOperator{\Dec}{Dec}
\newcommand\modelversion[2][en]{v$\,^\texttt{\;\;\;#1}_{\!#2}$}
\newcommand\gtt[1][\textvisiblespace]{\textcolor{darkgreen}{\texttt{#1}}}

\makeatletter
\def\thickhline{%
  \noalign{\ifnum0=`}\fi\hrule \@height \thickarrayrulewidth \futurelet
   \reserved@a\@xthickhline}
\def\@xthickhline{\ifx\reserved@a\thickhline
               \vskip\doublerulesep
               \vskip-\thickarrayrulewidth
             \fi
      \ifnum0=`{\fi}}
\makeatother

\renewenvironment{thebibliography}[1]{%
  \begin{oldthebibliography}{#1}%
    \setlength{\parskip}{0.33ex}%
    \setlength{\itemsep}{0.28ex}%
}%
{%
  \end{oldthebibliography}%
}

\newlength{\thickarrayrulewidth}
\usepackage{etoolbox}
\title{Online Gesture Recognition using Transformer\\ and Natural Language Processing}

\name{ 
  Guénolé C.M. Silvestre$\,^{\dagger}$,
  Félix Balado$\,^{\dagger}$, 
  Olumide Akinremi$\,^{\dagger}$ and
  Mirco Ramo$\,^{\dagger,\ddagger}$
  }
\address{
  $^{\dagger}$ School of Computer Science, University College Dublin, Ireland\\ 
  $^{\ddagger}$ Dip. Ingegneria dell'Informazione, University of Pisa, Italy
  }

\begin{document}
\ninept
\maketitle


\begin{abstract}
  The Transformer architecture is shown to provide a powerful machine transduction framework for online handwritten gestures corresponding to glyph strokes of natural language sentences.
  The attention mechanism is successfully used to create latent representations of an end-to-end encoder-decoder model, solving multi-level segmentation while also learning some language features and syntax rules.
  The additional use of a large decoding space with some learned Byte-Pair-Encoding (BPE) is shown to provide robustness to ablated inputs and syntax rules.
  The encoder stack was directly fed with spatio-temporal data tokens potentially forming an infinitely large input vocabulary, an approach that finds applications beyond that of this work.
  Encoder transfer learning capabilities is also demonstrated on several languages resulting in faster optimisation and shared parameters. 
  A new supervised dataset of online handwriting gestures suitable for generic handwriting recognition tasks was used to successfully train a small transformer model to an average normalised Levenshtein accuracy of 96\% on English or German sentences and 94\% in French.
\end{abstract}



\section{Introduction}
\label{sec:intro}

Handwriting Character Recognition (HCR) when associated with touch-sensitive display panels provides an intuitive and seamless input mechanism 
eschewing the need for structured UIs such as virtual keyboards, often slow and error-prone while also distant to the natural handwriting experience. 

In this context, online gesture recognition of glyphs refers to the problem of mapping a set of user gestures corresponding to sequences of spatio-temporal samples into their corresponding symbolic representation.
Each $n$-dimensional sample individuates a touch. A coherent and consecutive sequence of touches defines a stroke that can be combined to form glyphs. Glyphs correspond to characters or symbols encoded in a language vocabulary. 
It also extends to a wider context of symbols drawn from large alphabets composed of glyphs with many strokes such as logographic systems or mathematical expressions.
Table~\ref{table:terminology} formalises the terminology adopted in this work.


\begin{table}[t]
  \renewcommand{\arraystretch}{1.3}
  \caption{\label{table:terminology} Terminology}
  \setlength\tabcolsep{1ex}
  \small
  \begin{tabularx}{\linewidth}{p{0.12\linewidth}||>{\raggedright\arraybackslash}p{0.81\linewidth}}
    \hline
    \bf Term &  \bf Definition\\
    \hline\hline
    touch/ point &  $(x,y,[t, p])$ location on touch panel sampled at $t$ with finger pressure $p$\\\hline 
    stroke & Sequence of points where finger consecutively touches the panel \\\hline 
    glyph & List of one or more strokes individuating an element in the vocabulary\\\hline
    symbol & Item in a vocabulary of 57 symbols $\{ {\tt a}-{\tt z},$ ${\tt A}-{\tt Z},$ \textvisiblespace$, \token{unk}, \token{bos}, \token{eos}, \token{pad}\}$ or 2\,000 BPE tokens trained from $\{{\tt en}, {\tt fr}, {\tt de}\}$ datasets \\\hline
  \end{tabularx}
  \vskip -1em
\end{table}

These applications must jointly solve a number of tasks, namely 
i) accurate feature extraction in a multi-dimensional spatio-temporal space, 
ii) segmentation of stroke sequences to identify glyphs,
iii) glyph segmentation for the purpose of word/sentence recognition, and
iv) the encoding of syntax rules and language patterns to form a correct symbolic output.
An example of an online gesture sequence is depicted in Fig.~\ref{fig:online_gesture_example}.


\section{Related Work \& Contributions}\label{sec:related_work}
The field of HCR consists of techniques aiming at generating text directly from handwritten inputs. 
Most solutions rely on offline data due to dataset availability~\cite{Plamondon:2000,Sinwar:2021}.
However, the temporal dimension provides some valuable information that may simplify stroke segmentation, avoiding recourse to convoluted regression strategies such as text-line segmentation~\cite{Barakat:2021}. 
As a result, online methods generally exhibit superior performance over offline counterparts as reported in~\cite{Kim:2014,Shrivastava:2019,Corr:2017}.
With the growing popularity of the attention mechanism~\cite{Bahdanau:2014,Poulos:2021}, the field remains in constant development with much effort and resources devoted to improving existing techniques~\cite{Keysers:2017,Graves:2013}.
In a related sub-problem, Handwritten Mathematical Expression Recognition (HMER) consists in the generation of mathematical expressions using formal syntaxes, with state-of-the-art HMER models reaching impressive levels of accuracy, particularly when exploiting attention~\cite{Li:2020} and combining modalities from online and offline data~\cite{Wang:2019}.
However, these models fail to learn the intrinsic structure of expressions.
This is somewhat addressed in~\cite{Zhang:2021} using an RNN encoder along with a syntactic tree decoder.

In the context of sequence transduction tasks~\cite{Sutskever:2014}, the Transformer~\cite{Vaswani:2017} framework stands as the state-of-the-art on almost all NLP tasks~\cite{Devlin:2018,Brown:2020,Wolf:2020,Garg:2018}, eschewing recourse to recurrent or convolutional units. 
This powerful sequence mapping architecture relies entirely on the attention mechanism~\cite{Bahdanau:2014}, allowing for significant parallelisation and unattenuated gradient flow. 
It has also been successfully applied to a wider and more generic group of sequence transduction problems~\cite{Parmar:2018,Huang:2018,Zhao:2021,Kozlov:2020,DEusanio:2020}.
The Transformer popularity saw many proposals to revisit and optimise its design ~\cite{Wang:2020,Kitaev:2020,Choromansky:2020,Rao:2021} but very few are capable of clearly outperforming the original topology. 

This work follows the seminal work by~\cite{Vaswani:2017} and proposes to reformulate the online gesture recognition problem as a neural transduction task, leveraging the power of attention in the context of natural languages. 


\subsubsection*{Main Contributions}

\vspace*{0em}
\begin{enumerate}[(i)]\itemsep=0em
    \item New online datasets are proposed for handwritten text in natural languages (cf. Section~\ref{sec:datasets}) and suitable for a wide range of supervised and unsupervised machine learning applications.
    \item The attention mechanism is shown to successfully learn and represent implicit structures of spatio-temporal gesture data. 
    \item The power of transformers is demonstrated not only as language models but also as a solution to several sequence mapping tasks, with transfer learning behaviours observed for the encoder\footnote{Encoder with frozen parameters pre-trained on English word dataset can be used on other languages with alphabets using some potentially unknown glyphs.}.
    \item A small footprint\footnote{Despite its small size, model can perform the tasks of glyph segmentation, character recognition, word segmentation and sentence construction at remarkable performance levels, learning efficiently the input/output mapping.} topology is proposed as an end-to-end model, with fast optimisation, high accuracy and suitable for edge inference. 
    \item Model robustness is demonstrated on ablated inputs with the ability to generate grammatically compliant expressions in case of missing strokes. 
    \item Multi-level segmentation capability is highlighted in the correlation between syntactically correct predictions and their explainability in cross-attention visualisation.
\end{enumerate}


\section{Datasets}\label{sec:datasets}
An important contribution of this work is that of online gesture datasets for text in English ({\tt en}), French ({\tt fr}) or German ({\tt de}), suitable for investigating the task of HCR but also segmentation of touch, stroke or glyph, and eventually grammatical/syntactical compliance of expressions in natural languages.
Our handwritten database is presented as a coherent collection of tables composing a relational schema with spatio-temporal data for Roman alphabets, Arabic numerals~\cite{Corr:2017}, mathematical and punctuation symbols, collected from volunteers writing on touch panels.
This stage saw the contribution of over 600 subjects for a total of 69\,278 labelled glyphs composed by 93\,330 strokes, with over 2 million touches.
The dataset can be used at different levels of granularity, namely \textit{touch}, \textit{stroke} and \textit{glyph}.
In this work, we report results at the stroke level, leaving the burden of glyph segmentation to the model.

Subjects have been split into training, validation and test sets (60/20/20 proportions) such that models were tested on unseen handwriting styles to ensure accurate estimation of the generalisation power. 
When online HCR was performed in the context of natural languages, the WMT19 news dataset~\cite{WMT19:2019} was used to generate sequences of spatio-temporal gestures from our relational schema for all available sentences ($\simeq 100\,000$ sentences chosen to be no longer than 200 input strokes). 
Pre-processing was applied to substitute or omit characters where gesture data was unavailable e.g. {\tt \{é, è, ê\} $\rightarrow$ e} in {\tt fr} or {\tt ß $\rightarrow$  ss} in {\tt de}.
While these datasets are much smaller than those used to train large NLP models~\cite{Radford:2019}, it was found to be sufficient to demonstrate the model's ability to successful learn some language features at the sentence level.


\begin{figure}[t]
    \rule{\linewidth}{0.3pt} \\[0.3em]
    \includegraphics[width=\linewidth]{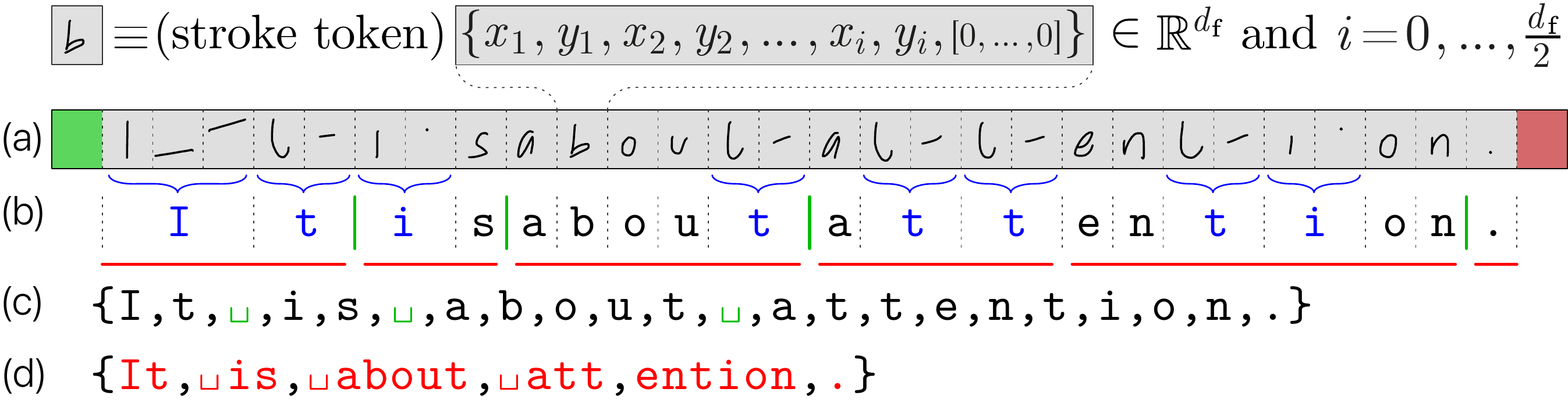} \\[-1.1em]
    \rule{\linewidth}{0.3pt}\\[-2em]
    \caption{\label{fig:online_gesture_example} 
    Online gesture example of an input stroke sequence (a) for the sentence (b) and its corresponding output list -- symbols in (c) and BPE tokens in (d). 
    Cells in (a) depict the linear interpolation of spatio-temporal points that forms input tokens. Green and red cells denote the \token{bos} and \token{eos} input token respectively. 
    Stroke sequence segmentation, glyph segmentation and parsing of language tokens are colour coded with with blue, green and
    red, respectively.
    }
  \end{figure}


\section{Transformer Architecture \& Experimental Details}
\label{sec:transformer}
Our model~\cite{Olu:2021} leverages the original Transformer~\cite{Vaswani:2017} architecture. However crucial modifications are introduced to work with spatio-temporal data. 
Given some input sequence, $X\in \mathbb{R}^{\df \times n}$, of $n$ stroke tokens defined as interleaved spatio-temporal data with zero-padding of fixed-length $\df$ ---appropriately prefixed, suffixed and padded at the end with \token{eos}, \token{bos} and \token{pad} tokens respectively ---, a mask $M_x$ is computed to ensure encoder's attention is only paid to valid online data tokens.
Here, $\df=64 \times 2$ allowing for a maximum of 64 $(x,y)$ touch samples per stroke. 

As the input is composed of spatio-temporal information corresponding to touches\footnote{Explicit sampling time and pressure $(t, p)$ features are observed to have no impact on performance suggesting some level of information redundancy.}, each encoder token embeds a stroke as $\df$ scalars (cf. Fig.~\ref{fig:online_gesture_example}) resulting in the identification within a potentially unbounded input vocabulary and therefore eschewing any form of embedding.

\vskip 0.5em\noindent\textbf{Positional encoding} provides a strategy to embed the positional information of input tokens in the encoder, a necessary operation since the attention mechanism has no built-in concept of sequentiality. 
Frequency modulation is proposed in~\cite{Vaswani:2017}. 
However, since we observed no performance gain with such a strategy, we use a learnable 1D embedding based on the token index. Stroke positions are encoded in $P_x\in \mathbb{R}^{\df \times n}$.

\vskip 0.5em\noindent\textbf{The encoder} is trained to learn some latent sequence representation $Z=\Enc(X + \alpha P_x, M_x) \in \mathbb{R}^{\da \times \dh \times n}$ where $\alpha$ is a scaling factor blending input data and positional information,  $\da$ the number of attention heads (2 to 4) and $\dh$ the hidden state dimension of the attention heads.
The encoder consists of a stack of $\le$ identical multi-head vanilla self-attention layers and a positional feed-forward network of hidden dimension $\dpf$. Each layer is followed by a residual connection before layer-normalisation.

When exploring transfer learning capabilities, an encoder with $\Thetae\!=\!523\,520$ parameters is used as a feature extractor resulting in a considerable speed-up during training and model optimisation.

\vskip 0.5em\noindent\textbf{The decoder} generates a causal sequence of tokens in an auto-regressive manner given some vocabulary and relative token encoding. 
It is initialised with the \token{bos} token and iteratively outputs a new token using greedy sampling of the decoder's softmax output until the \token{eos} token is predicted or the maximum sequence length, $m$, is reached.
The decoder also consists of $\ld$ identical layers each composed by: i) a masked self-attention layer (with $\da$ attention heads of hidden state dimension $\dh$) that prevents the decoder from peeking at the subsequent tokens, ii) a cross-attention layer that attends over the encoder output $Z$ to generate predictions, and iii) a feed-forward layer just as in the encoder but of dimension $k\dpf$ where $k\!=\!1$ to $3$.

At each step, the decoder's input, $Y_{i-1}$, is an auto-regressive sequence of tokens adequately masked with $M_y$ and used to predict the next token of the output sequence: $Y_i=\Dec(Y_{i-1}, M_y, Z)$.
All $\Thetad$ parameters of the decoder were trained from some randomly initialised state.

\vskip 0.5em\noindent\textbf{Experimental details:} 
All models were configured with $\df\!=\!\da\times\dh\!=\!\dpf\!=\!128$. For \modelversion[rnd]{61}--\modelversion[gow]{68}, $n\!=\!2 m\!=\!48$ and $k\!=\!1$. For \modelversion[en]{74}--\modelversion[de]{93}, $n\!=\!2 m\!=\!200$ and $k\!=\!3$. Parameters of the attention layer stack are detailed Tables~\ref{table:hyperparams_search} and \ref{table:language_models}.
Models were trained on Nvidia TitanX GPUs\footnote{Nvidia is acknowledged for the donation of GPUs}, for a maximum of 5000 epochs, a 256 batch size, using cross-entropy loss and Adam optimiser with a decay schedule (initial learning rate of $8\times 10^{-4}$ and halving every 30 epochs).


\section{Experimental Results}
\label{sec:results}
A series of experiments were carried out to determine suitable model hyper-parameters and investigate the benefits of increasing the decoder's output dimension using vocabularies (vocabs) of various sizes: i) a small vocab of 57 alphabetic symbols, and ii) larger vocabs with 2\,000 Byte-Pair-Encoding (BPE) symbols trained on three natural languages ({\tt en}, {\tt fr}, {\tt de}).
Models were evaluated using a number of performance metrics on the test sets and results are reported in terms of Cross-Entropy Loss (XEL) and two edit distance metrics, namely: normalised Levenshtein distance~\cite{Yujian:2007} Accuracy (LA) and Character Error Rate (CER).

Hyper-parameters were determined from XEL results obtained on models trained with pseudorandom letter words as reported in Table~\ref{table:hyperparams_search}. 
Since this only required to solve the tasks of glyph segmentation and classification, splitting hidden states in two multi-attention heads was sufficient.
A five-layer encoder abstraction performed best, a result consistent with previous observations in an online classification context using convolutional topologies~\cite{Corr:2017}.
Note that there are no benefits in increasing the decoder's layer abstraction since output tokens are i.i.d. in a small 57 symbol vocab, with no symbol patterns to be exploited.


\begin{table}[th]
    \vspace*{-0.7em}
    \caption{\label{table:hyperparams_search} Hyper-parameter search for models trained on stroke sequence corresponding to pseudorandom letter words. \scalebox{0.96}[1]{Performance is reported in terms of Cross-Entropy Loss (XEL).}}
    \begin{center}
    \def\arraystretch{1.2}
    \vspace*{-0.8em}
    \small
        \begin{tabularx}{\linewidth}{c|| c c c c} \hline
        Name & Enc ($\le, \da$) & Dec ($\ld, \da$) & $\Thetae+ \Thetad$ & XEL \\ \hline \hline
        \modelversion[rnd]{61} & 2L, 4H & 2L, 4H & 554\,552 & 0.761  \\
        \modelversion[rnd]{62} & 3L, 2H & 2L, 2H & 654\,136 & 0.881  \\
        \modelversion[rnd]{63} & 4L, 4H & 4L, 4H & 1\,085\,496 & 0.914  \\
        \modelversion[rnd]{64} & \textbf{5L}, \textbf{2H} & \textbf{2L}, \textbf{2H} & 850\,232 & \textbf{0.690}  \\
        \modelversion[rnd]{65} & 5L, 2H & 4L, 2H & 1\,185\,080 & 1.556  \\
        \modelversion[rnd]{66} & 7L, 2H & 2L, 2H & 1\,052\,472  & 0.895  \\
        \modelversion[rnd]{67} & 7L, 2H & 4L, 2H & 1\,384\,248 & 1.332  \\\hline
        \end{tabularx}
    \end{center}
    \vspace*{0.2em}
\end{table}


\begin{table}[t]
    \caption{\label{table:language_models} Model performance reported in term of Cross-Entropy Loss (XEL), normalised Levenshtein distance Accuracy (LA) and Character Error Rate (CER). \modelversion[gow]{68} trained on a group of {\tt en} words (57-symbol vocab), \modelversion{80}--\modelversion[de]{93} trained on sentences in three languages (2\,000 BPE vocab). $\Thetae=523\,520$ (all), $\Thetad=330\,041$ (\modelversion[gow]{68}) and $1\,453\,520$ (others). Fine-tuned models provide best performance. }
    \centering
    \bgroup
    \def\arraystretch{1.2}
    \setlength\tabcolsep{1.3ex}
    \vspace*{0.3em}
    \small
    \begin{tabularx}{\linewidth}{c || c c c c c}
      \hline
      Name &  Enc ($\le, \da$) & Dec ($\ld, \da$) & XEL & LA(\%) & CER \\ \hline \hline
      \modelversion[gow]{68} & 5L, 4H$^\ddagger$ & 2L, 4H$^\ddagger$ & 0.282 & 91.07 & 0.089 \\ \hline
      \modelversion{74} & 5L, 4H$^\ddagger$ & 4L, 4H$^\ddagger$ & 0.260 & 94.04 & 0.066 \\ \hline
      \modelversion{80} & 5L, 4H$^\ast$ & 4L, 4H$^\ddagger$ & 0.183 & \textbf{95.90} & 0.045 \\ 
      \modelversion[fr]{82} & 5L, 4H$^\ast$ & 4L, 4H$^\ddagger$ & 0.313 & 93.21 & 0.074 \\ 
      \modelversion[de]{83} & 5L, 4H$^\ast$ & 4L, 4H$^\ddagger$ & 0.285 & \textbf{94.79} & 0.057 \\ \hline
      \modelversion[fr]{92} &  5L, 4H$^\dagger$ & 4L, 4H$^\dagger$ & 0.293 &\textbf{93.51} & 0.071 \\ 
      \modelversion[de]{93} & 5L, 4H$^\dagger$ & 4L, 4H$^\dagger$ & 0.298 & 94.52 & 0.060 \\ \hline 
    \end{tabularx}
    \egroup
    \linebreak \scriptsize{\rule{1.5em}{0em}$^\ddagger$\scalebox{.88}[1.0]{Trained from scratch from some randomly initialised state}}\hfill~
    \linebreak \scriptsize{\rule{1.5em}{0em}$^\dagger$\scalebox{.88}[1.0]{Fined-tuned parameters using transfer learning of \modelversion{74} encoder}}\hfill~
    \linebreak \scriptsize{\rule{1.5em}{0em}$^\ast$\scalebox{.88}[1.0]{Frozen parameters using transfer learning of \modelversion{74} encoder}}\hfill~
    \vspace*{-0.5em}
  \end{table}

Table~\ref{table:language_models} summarises the main results with hyper-parameter configuration and performance evaluation for models trained on natural languages. 
Since this requires solving multilevel segmentation tasks, the hidden states of a 4-layer decoder abstraction were split into 4 attention heads.
Model \modelversion[en]{68} was trained on short {\tt en} sentence chunks to output symbols using the 57 symbol vocab, with some modest data augmentation strategies using affine transformations. 
Despite its smaller size, model demonstrates strong end-to-end recognition capabilities, learning some syntax rules.
Larger models ($\Thetae\!+\!\Thetad\!=\!\!1.9\,$M) were subsequently trained on full sentences in different languages.
\modelversion[en]{74} was optimised from a random state with the same output configuration, exhibiting some improved performance over the smaller \modelversion[gow]{68} model. 
Its encoder parameters were then used as a feature extractor when training larger models outputting BPE tokens ({\tt en}, {\tt fr}, {\tt de}) with a significant optimisation speed-up.
With a performance reaching a LA of 96\% \hstretch{1}{\textrm{(4.5\% CER, 14.7\% WER), the {\tt en} model's}} improvement is attributed to the larger decoder output dimension that provides some error correction mechanisms as confirmed in the robustness analysis of Table~\ref{table:ablation_studies}, eliminating some spelling errors.
Although the encoder was optimised on inputs associated with {\tt en} words, the {\tt de} model \modelversion[de]{83} fell just 0.4\% short of the {\tt en} model accuracy, with strong performance also observed with the {\tt fr} model \modelversion[fr]{82} tailing by 1.5\%. 
It is worth noting that fine-tuning of the {\tt fr} model over a couple of epochs resulted in modest performance gains.

\subsubsection*{Robustness and Visualisation} 
\label{ssec:robustness_and_visualisation}
Model robustness is investigated with input stroke ablation and deliberate erroneous input while observing the model's ability to enforce syntax rules and correct spelling.

As all dataset expressions end with some punctuation mark, the learning of this rule is observed in all models as shown in Table~\ref{table:ablation_studies}. 
In addition, interrogative forms are detected from the subject/verb inversion, with correct insertion of a question mark as shown in the {\tt fr} example\:\raisebox{-0.2em}{\circled{\tiny\bf 2}}. 
One can also note the correct hyphenation token inferred between verb and subject pronoun as expected in {\tt fr} syntax, along with spacing before the question mark.
The large BPE vocab also provides some robustness to spelling errors as shown in the {\tt de} example\:\raisebox{-0.2em}{\circled{\tiny\bf 3}}. 
This was not observed to the same extent on models with smaller output vocab such as \modelversion{68}. 
Finally, contraction of {\tt en} verbs is correctly detected and expanded as seen in the {\tt en} example\:\raisebox{-0.2em}{\circled{\tiny\bf 1}}. 
These observations demonstrate that the models are capable of learning non-trivial valuable syntax and grammar rules along with some language features despite the small dataset and model size.


\begin{table}[t]
    \caption{Robustness to elided input, spelling errors, and mis\-sing punctuation. BPE models can infer correct expressions.}
    \label{table:ablation_studies}
    \setlength\tabcolsep{0.5ex}
    \def\arraystretch{1.7}
    \vspace*{0.3em}
    \small
    \begin{tabularx}{\linewidth}{p{\linewidth} } \hline
        \scalebox{1}[1.0]{Input ($X$) for glyph symbols ($S$) and inference output ($\hat{Y}$)} \\ 
        \thickhline
        {$X\!\!=$\,\raisebox{-0.32em}{\includegraphics[height=1.2em]{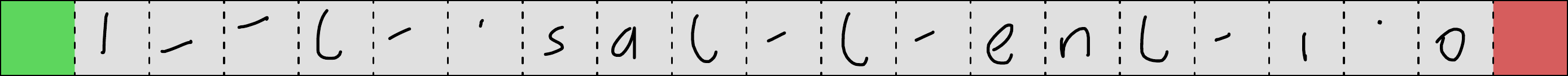}}}\\
        {$S\!=\!\left\{\texttt{I,t,',s,a,t,t,e,n,t,i,o}\right\}$}\\[-0.5em]
        {\modelversion[en]{68}
        $\rightarrow \hat{Y}\!\!=\!\left\{\texttt{I,t,\gtt{},\gtt[i],s,\gtt{},a,t,t,e,n,t,i,o,\gtt[n]}\right\}$}\\[-0.5em]
        {\modelversion[en]{80}
        $\rightarrow \hat{Y}\!\!=\!\left\{\texttt{It,\gtt{}\gtt[i]s,\gtt{}att,entio\gtt[n],\gtt[.]}\right\}$}
        \hfill \circled{\tiny\bf 1}\rule{0.5ex}{0em}\\
        \hline
        {$X\!\!=$\,\raisebox{-0.32em}{\includegraphics[height=1.2em]{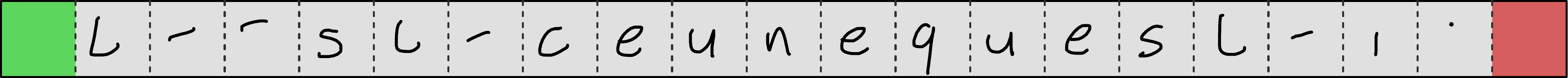}}}\\
        {$S\!=\!\left\{\texttt{E,s,t,c,e,u,n,e,q,u,e,s,t,i}\right\}$}\\[-0.5em]
        {\modelversion[fr]{82}
        $\rightarrow \hat{Y}\!\!=\!\left\{\texttt{Est,\gtt[-],ce,\gtt{}une,\gtt{}questi\gtt[on],\gtt{}\gtt[?]}\right\}$}
        \hfill \circled{\tiny\bf 2}\rule{0.5ex}{0em}\\
        \hline
        {$X\!\!=$\,\raisebox{-1.6em}{\includegraphics[height=2.4em]{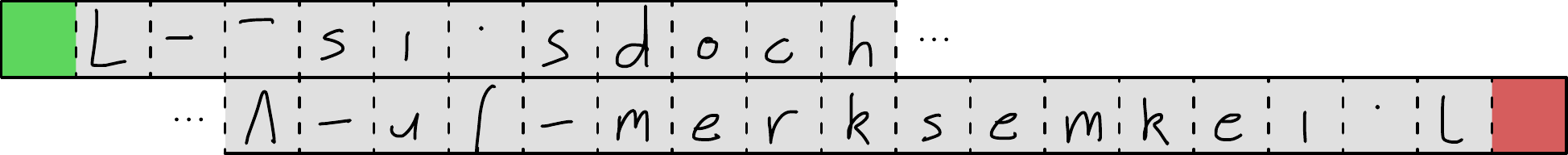}}}\\
        {$S\!=\!\left\{\scalebox{1}[1.0]{\texttt{E,s,i,s,d,o,c,h,A,u,f,m,e,r,k,s,e,m,k,e,i,t}}\right\}$}\\[-0.5em]
        {\modelversion[de]{83}
        $\rightarrow \hat{Y}\!\!=\!\left\{\scalebox{1}[1.0]{\texttt{Es,\gtt{}is\gtt[t],\gtt{}doch,\gtt{}Auf,mer,k,s\gtt[a]m,keit,\gtt[.]}}\right\}$}
        \hfill \circled{\tiny\bf 3}\rule{0.5ex}{0em}\\
        \hline
    \end{tabularx}
\end{table}

\begin{figure}[tbp]
    \centering
    \includegraphics[width=\linewidth]{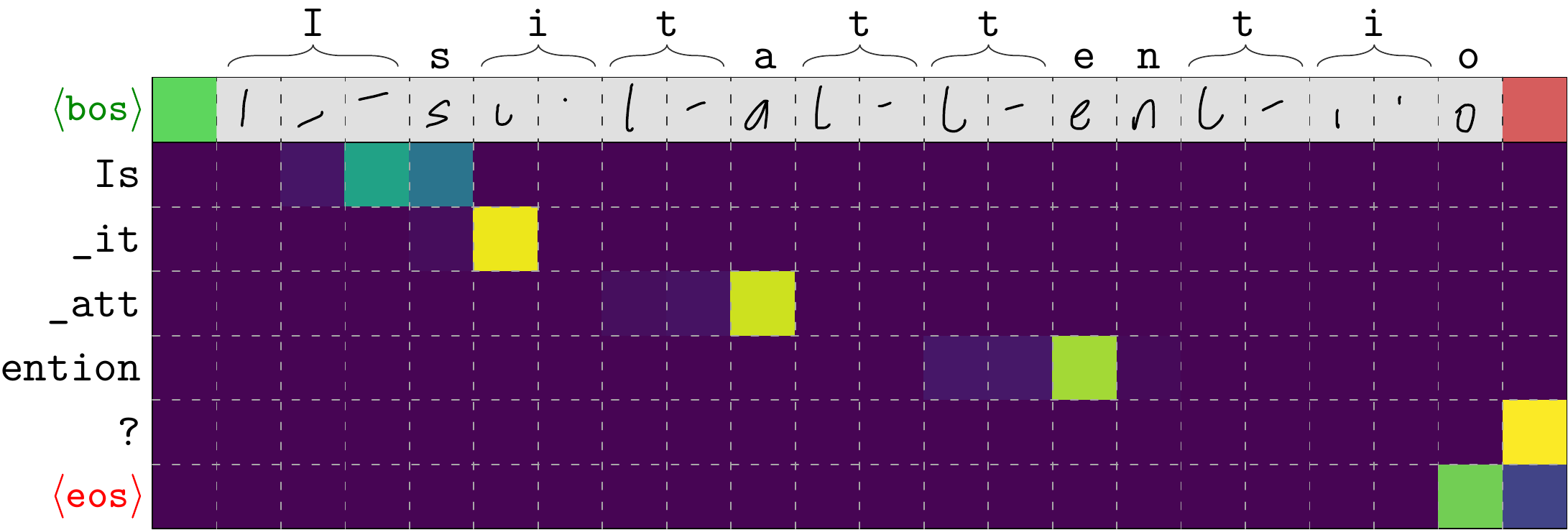}
    \vskip -1em
    \caption{\label{fig:attention_plot} Cross-attention plot of head 3--layer 4 for model \modelversion{80} showing output tokens tracking its first stroke. Question mark token is focusing on input \raisebox{-0.1em}{\textcolor{redish}{\rule{0.8em}{0.8em}}} token.} 
\end{figure}

Visualisation of the attention mechanisms provides some interesting insights in the learning process.
Fig.~\ref{fig:attention_plot} depicts weights of the decoder's cross-attention over the encoder's output. 
It shows that head 3 of layer 4 is key to token segmentation with strong attention paid to the first token stroke solving the word segmentation task. 
The large BPE space is also leveraged for the auto-completion of the final text token, while insertion of the punctuation mark is carried out by tracking the final stroke.
The \token{eos} token output is seen focusing on both final stroke and \token{eos} stroke input.


\section{Conclusion}
This work posed the problem of online HCR as a transduction task, using a Transformer framework to learn the complex mapping of online input gesture data corresponding to handwritten strokes of natural language sentences. 
The encoder's input was modified to receive spatio-temporal data as real-valued tokens, operating at stroke level without the need for mapping on a fixed input vocabulary.
Models were shown to predict text with very high accuracy by handling internally the multi-level segmentation of inputs (at glyph and word levels), and also understanding and learning how to represent and enforce syntactic and semantic rules of data.
Index positional encoding was shown to be as effective as cosine modulation yet standing as a simpler and more natural encoding for the position information. 
Although the Transformer's ability to generate complex representations and learn non-trivial input/output mapping between sequences is well established~\cite{Devlin:2018,Parmar:2018}, the challenge was further pushed in this work in the absence of ad hoc syntax, semantic rules, or engineered loss computation and architecture.

In addition, an encoder trained on a specific language was successfully used as a frozen feature extractor in the optimisation of decoders in several other language domains. 
Such transfer learning capabilities suggest that pre-trained encoders can create general latent representations suitable for problems of different nature, resulting in model size reduction, training acceleration with no need for fine-tuning or explicit domain adaptation. 
This will also benefit applications where computational power/time and dataset size are limited.

The objective of this work was not so much to push out some state-of-the-art model but rather to state some important considerations that may be the starting points for future works on sequentiality of other signal types. 
Neural transduction may be extended in this way to online data at different granularity levels, with no need for separate input segmentation or complex positional embeddings.
With larger language datasets and computational resources, this approach may reveal deeper language modelling capabilities to similar levels observed in BERT or GPT~\cite{Devlin:2018,Radford:2019}, and this straight from digital signals.
The end-to-end encoder-decoder models in this work achieved a normalised Levenshtein accuracy of 94\% to 96\% at sentence levels on the three languages considered and directly from online data.


\bibliographystyle{IEEEbib}
\bibliography{refs}

\end{document}